%% file: main.tex
\documentclass[10pt,twocolumn,letterpaper]{article}

\usepackage{cvpr}              % To produce the CAMERA-READY version
\usepackage[accsupp]{axessibility} 
\usepackage{svg}
\usepackage{multirow}
\usepackage{pifont}% http://ctan.org/pkg/pifont
\newcommand{\cmark}{\ding{51}}%
\newcommand{\xmark}{\ding{55}}%
\input{preamble}
\definecolor{cvprblue}{rgb}{0.21,0.49,0.74}
\usepackage[pagebackref,breaklinks,colorlinks,allcolors=cvprblue]{hyperref}
\usepackage{cuted}

\def\paperID{***} % *** Enter the Paper ID here
\def\confName{CVPR}
\def\confYear{2026}

\title{Calibri: Enhancing Diffusion Transformers via Parameter-Efficient Calibration}

\author{Danil Tokhchukov$^*$\\
MSU\\
\and
Aysel Mirzoeva\\
MSU\\
\and
Andrey Kuznetsov\\
FusionBrain Lab, AXXX\\
\and
Konstantin Sobolev$^{*\dagger}$\\
FusionBrain Lab, AXXX\\
MSU\\
}

\begin{document}
\maketitle

\def\thefootnote{*}\footnotetext{The authors contributed equally.}
\def\thefootnote{$\dagger$}\footnotetext{Project lead, correspondence: ksobolev.info@gmail.com}
\def\thefootnote{$\ddagger$}\footnotetext{Project: \url{https://v-gen-ai.github.io/Calibri-page/}}

\begin{strip}
\vspace{-5.4em}
  \centering
  \includegraphics[width=\textwidth,trim={-15em 1em -15em 6em},clip]{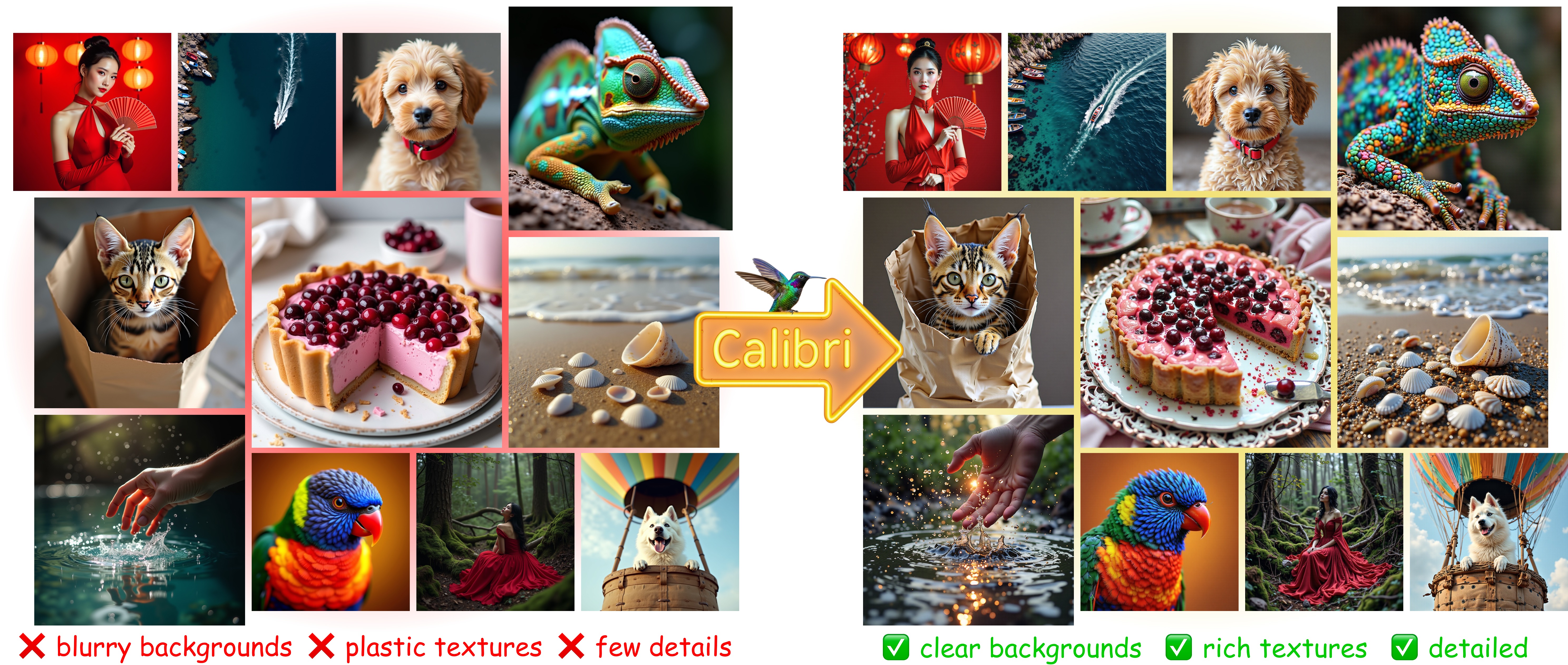}

  \captionof{figure}{Introducing \textit{Calibri} -- a parameter-efficient method for diffusion transformer alignment. By optimizing only $\sim 10^2$ parameters, \textit{Calibri} significantly enhances the model's generation quality.}
  \label{fig:visual_abs}
\end{strip}

% \begin{center*}
%   \includegraphics[width=0.95\textwidth,trim={0em 1em 0em 6em},clip]{figures/graphic_abstract__v4.jpg}
%   \captionof{figure}{Introducing \textit{Calibri} -- a parameter-efficient method for diffusion transformer alignment. By optimizing only $\sim 10^2$ parameters, \textit{Calibri} significantly enhances the model's generation quality.}
%   \label{fig:visual_abs}
% \end{center*}

\input{sec/0_abstract}    
\input{sec/1_intro}

\input{sec/2_related}

\input{sec/3_method}

\input{sec/4_experiments}
\input{sec/5_conclusion}
{
    \small
    \bibliographystyle{ieeenat_fullname}
    \bibliography{main}
}


\appendix
\input{sec/X_suppl}

\end{document}

%% file: sec/0_abstract.tex
\begin{abstract}

% In this paper, we uncover the hidden potential of Diffusion Transformers (DiTs) to significantly enhance generative tasks. By analyzing the denoising process, we demonstrate that the performance of DiT blocks can be improved using a single learned scaling parameter. Based on this insight, we propose \textbf{Calibri}, a parameter-efficient method that optimally calibrates DiT components to boost generative quality. \textbf{Calibri} formulates calibration as a black-box reward optimization problem, solved efficiently via an evolutionary algorithm, and modifies only $\sim 10^2$ parameters. Moreover, \textbf{Calibri} introduces a novel inference-time ensemble scaling strategy, further enhancing generative performance. Experimental results show that despite its lightweight approach, \textbf{Calibri} achieves consistent improvements across diverse text-to-image models. Interestingly, \textbf{Calibri} also reduces the number of inference steps required for image generation while maintaining high quality.

In this paper, we uncover the hidden potential of Diffusion Transformers (DiTs) to significantly enhance generative tasks. Through an in-depth analysis of the denoising process, we demonstrate that introducing a single learned scaling parameter can significantly improve the performance of DiT blocks. Building on this insight, we propose \textbf{Calibri}, a parameter-efficient approach that optimally calibrates DiT components to elevate generative quality. \textbf{Calibri} frames DiT calibration as a black-box reward optimization problem, which is efficiently solved using an evolutionary algorithm and modifies just $\sim 10^2$ parameters. 
% Additionally, we introduce \textbf{Calibri Ensemble} introduces an innovative inference-time ensemble scaling strategy to further boost generative performance. 
Experimental results reveal that despite its lightweight design, \textbf{Calibri} consistently improves performance across various text-to-image models. Notably, \textbf{Calibri} also reduces the inference steps required for image generation, all while maintaining high-quality outputs.

\end{abstract}

%% file: sec/1_intro.tex
\section{Introduction}
\label{sec:intro}

In recent years, the field of visual content generation has experienced significant advancements, largely fueled by the development of diffusion models~\cite{ddpm, rombach2022high}. Cutting-edge models like Stable Diffusion 3~\cite{esser2024scaling} and FLUX~\cite{flux2024} have redefined the landscape of modern generative frameworks. These models represent a shift from the traditional UNet architecture~\cite{ronneberger2015u} to the more advanced Diffusion Transformer (DiT)~\cite{peebles2023scalable}, while also incorporating innovative techniques such as flow matching~\cite{lipmanflow} to enhance their capabilities. This powerful combination of a DiT backbone and flow matching has become the new de facto standard, extending far beyond text-to-image synthesis to power diverse tasks such as instruction-guided image editing~\cite{labs2025flux, wu2025qwen} and video generation~\cite{wan2025wan}.

Diffusion transformers are built from a sequence of identical blocks, each containing attention and MLP layers. Despite this uniform architecture, recent work suggests their functional contributions are highly uneven. For instance, Stable Flow~\cite{avrahami2025stable} identified "vital layers" within the transformer, whose exclusion from generation process significantly alters the model's output. This finding implies that not all layers contribute equally to the final generation.

% Diffusion transformers feature a relatively straightforward design - a sequence of identical blocks comprising attention layers and MLPs, modulated by temporal embeddings. Despite this straightforward design, recent findings from Stable Flow~\cite{avrahami2025stable} revealed that the contribution of individual layers within diffusion transformers is far from uniform. Specifically, their study identified the presence of vital layers, whose exclusion from generation process significantly alters the model's output.

% Inspired by this discovery, we analyze the contributions of individual DiT blocks to the output quality. Our analysis revealed two intriguing results: First, excluding certain blocks improves the quality of the generation. Second, for each block, an individual coefficient can be selected which, when multiplied, improves the quality of the model compared to the original. In other words, simple multiplication of each block output by single number can increase generation quality. Based on this observation, we hypothesize that diffusion transformers need calibration.

Building on this insight, we analyze the contribution of individual DiT blocks and uncover two surprising results. First, we find that selectively disabling certain blocks can actually improve generation quality, suggesting some may introduce detrimental artifacts. Second, we discover that a simple re-weighting of each block's output -- by multiplying it with a single learned scalar -- consistently enhances the model's performance over the original. These observations lead us to our central hypothesis: \textbf{\textit{The standard DiT architecture is sub-optimally weighted, and its performance can be significantly improved through a simple post-hoc calibration of its blocks.}}

Motivated by this hypothesis, we propose \textit{Calibri}, a parameter-efficient approach designed to calibrate the contributions of DiT's architectural components and improve generation quality (Figure~\ref{fig:visual_abs}). Specifically, we frame the process of determining calibration coefficients as a black-box optimization problem with only $\sim 10^2$ parameters. The objective is to maximize the quality of model outputs, as measured by a reward model~\cite{ma2025hpsv3, wang2025unified}. To solve this optimization problem, we leverage the gradient-free evolutionary strategy CMA-ES~\cite{hansen2003reducing, hansen2019pycma}, which effectively identifies optimal scaling coefficients. Furthermore, we introduce \textit{Calibri Ensemble}, which integrates multiple calibrated models to further boost generative performance. Notably, \textit{Calibri} also reduces the number of inference steps required for image generation, significantly improving both efficiency and quality. Extensive experiments across diverse baseline models validate the effectiveness of \textit{Calibri} in achieving consistent performance gains without computational overhead.

% In summary, our key contributions are as follows:

% \begin{enumerate}
%     \item We analyze the potential of DiT architectures and reveal that their generation quality can be significantly improved through simple block scaling;
%     \item We propose a parameter-efficient framework, \textit{Calibri}, which calibrates the contributions of DiT components to enhance denoising and generation quality within diffusion and flow-based models;
%     \item We present \textit{Calibri Ensemble}, a novel inference-time ensemble strategy that integrates multiple calibrated \textit{Calibri} models, leading to further improvements in generation quality;
%     \item Both \textit{Calibri} and \textit{Calibri Ensemble} are designed to seamlessly integrate with existing diffusion and flow-based models, offering a flexible and effective solution. We validate their performance by demonstrating substantial improvements in sample quality across diverse baseline methods.
% \end{enumerate}

%% file: sec/2_related.tex
\section{Related Work}
\label{sec:related}

\textbf{Diffusion Models Backbones.} Early diffusion models predominantly utilized U-Net~\cite{ronneberger2015u} backbones with residual blocks~\cite{he2016deep}, pixelwise self-attention~\cite{vaswani2017attention}, and cross-attention layers for text-image conditioning~\cite{ho2020denoising, rombach2022high, ho2022video, blattmann2023align}. Recently, the field has shifted towards Diffusion Transformer (DiT)-based architecture~\cite{peebles2023scalable}, which received significant attention due to the scalability of transformer models~\cite{vaswani2017attention}. One notable development is PixArt-alpha~\cite{chenpixart}, which effectively applied DiT for text-conditional generation while preserving the conventional cross-attention mechanism for text-based conditioning. using a conventional cross-attention mechanism for text conditioning. A key milestone in this evolution is the introduction of the Multimodal Diffusion Transformer (MM-DiT)~\cite{esser2024scaling}, which employs distinct transformers to process textual and visual inputs, subsequently combining their sequences through unified attention operations.

% \noindent
\textbf{Diffusion Model Backbone Interpretability.} 
Recent research has significantly advanced the understanding of diffusion model architecture, enabling novel applications. Early studies showed that cross-attention maps between text prompts and visual tokens produce high-quality saliency maps to predict spatial locations of textual concepts~\cite{tang2023daam}, applied in tasks like image editing~\cite{hertz2022prompt} and layout control~\cite{chen2024training, epstein2023diffusion}. Other works explored diffusion model components: Free-U~\cite{si2024freeu} highlighted the U-Net backbone’s denoising role and skip connections’ contribution of high-frequency features, improving its denoising efficacy. Additionally, methods like Stable Flow~\cite{avrahami2025stable} and FreeFlux~\cite{wei2025freeflux} analyzed Diffusion Transformer (DiT) blocks, identifying critical layers for image formation and differentiating positional versus content-focused layers, leading to training-free image editing techniques that leverage interpretability in diffusion models.
% Recent research has made significant strides in interpreting the architecture of diffusion models, uncovering insights into their inner workings and facilitating novel applications. Early studies demonstrated that cross-attention maps between text prompts and visual tokens can generate high-fidelity saliency maps, which accurately predict the spatial locations of textual concepts~\cite{tang2023daam}. These maps have found widespread applications in tasks such as image editing~\cite{hertz2022prompt} and layout control~\cite{chen2024training, epstein2023diffusion}. Another stream of work has focused on understanding the individual components within the diffusion model backbone. For instance, Free-U~\cite{si2024freeu} revealed that the primary backbone of U-Net is predominantly responsible for denoising, whereas skip connections primarily introduce high-frequency features to the decoder module. By strategically reweighting the contributions of these components, the authors improved the denoising efficacy of U-Net. Furthermore, approaches like Stable Flow~\cite{avrahami2025stable} and FreeFlux~\cite{wei2025freeflux} have investigated the roles of individual DiT (Diffusion Transformer) blocks during the generation process. Stable Flow identified "vital layers" critical to image formation, while FreeFlux uncovered layers that emphasize positional information versus those that focus on content. Both methods leveraged these insights to propose novel, training-free image editing techniques, showcasing the interpretability-driven advancements in diffusion models.

% \noindent
\textbf{Visual Generative Model Alignment.} Aligning generative models, including diffusion models and rectified flows, with human feedback has significantly improved their performance. Conventional methods rely on reward models to capture human preferences~\cite{ma2025hpsv3, xu2023imagereward, kirstain2023pick, wang2025unified} and often use RLHF-inspired techniques like reward backpropagation~\cite{clarkdirectly, xu2023imagereward}, Direct Preference Optimization (DPO)~\cite{wallace2024diffusion}, Differentiable Diffusion Preference Optimization (DDPO)~\cite{blacktraining}, and Group Relative Policy Optimization (GRPO)~\cite{liu2025flow}, which typically require full model fine-tuning, making them computationally expensive.

% \textit{In contrast, \textit{Calibri} uses a gradient-free evolutionary strategy to align diffusion and rectified-flow models with reward outputs, optimizing only $\sim 10^2$ parameters to efficiently match human preferences.}

%% file: sec/3_method.tex
\section{Method}
\label{sec:method}

\subsection{Preliminaries}

\paragraph{Diffusion transformer architecture}

\begin{figure}[t]
\begin{center}

\begin{subfigure}{.49\linewidth}

\includegraphics[width=\linewidth,trim={2em 1em 2em 2em},clip]{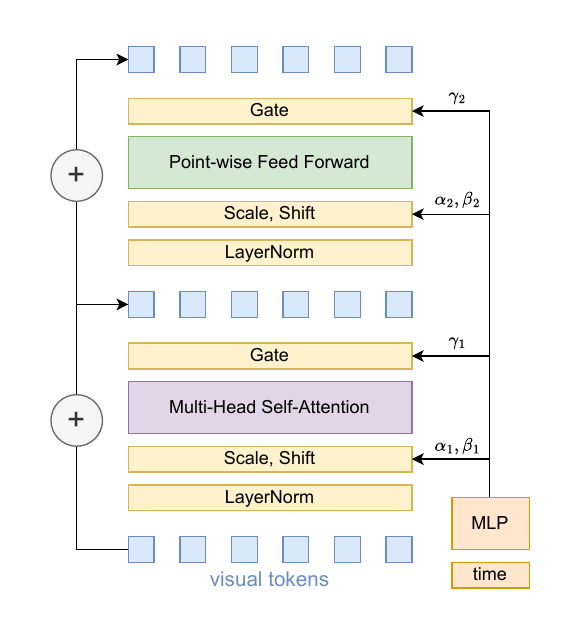}
\caption{DiT block scheme}
\label{subfig:dit_block_scheme}
\end{subfigure}
\hfill
\begin{subfigure}{.49\linewidth}

\includegraphics[width=\linewidth,trim={2em 1em 2em 2em},clip]{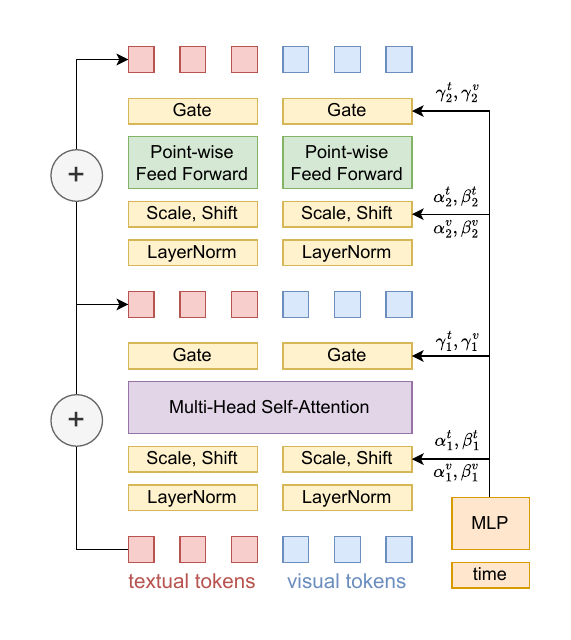}
\caption{MM-DiT block scheme}
\label{subfig:mm_dit_block_scheme}
\end{subfigure}

\caption{Illustration of DiT architectural components.}
\label{fig:dit_scheme}

\end{center}
\end{figure}

The Diffusion Transformer (DiT) architecture comprises sequential DiT blocks that transform input tokens into output tokens. Two main types of DiT blocks have been introduced:

\textbf{Stantdard DiT block}~\cite{peebles2023scalable} consists of Multi-Head Self-Attention (MHSA) layers and feed-forward layers, as illustrated in Figure~\ref{subfig:dit_block_scheme}. Both layers apply LayerNorm to the incoming data and are modulated by a time embedding. The modulation is achieved using vectors $\alpha, \beta, \gamma$, which are generated by a distinct Multi-Layer Perceptron (MLP). Output of layers can be described by this formula:

    \begin{equation} \label{eq:dit_layers}
    \begin{split}
        &x_l = x_{l-1} + \gamma_1 MHSA(\alpha_1 LN(x_{l-1}) + \beta_1), \\
        &x_{l+1} = x_{l} + \gamma_2 FF(\alpha_2 LN(x_l) + \beta_2),
    \end{split}
    \end{equation}

\noindent
where $x_{l-1}$ denotes the input token sequence, and $x_l$ and $x_{l+1}$ represent the intermediate and final outputs.

\textbf{MM-DiT block}~\cite{esser2024scaling} builds upon the structure of the Standard DiT block while introducing functionality for multimodal data processing. Specifically, MM-DiT combines textual and visual tokens via concatenation and processes them in parallel. Inter-modal communication is restricted to the MultiModal Attention Layer, enabling effective interaction between the two modalities. Separate modulation vectors are employed for each modality, denoted as $\alpha^v, \beta^v, \gamma^v$ for visual tokens and $\alpha^t, \beta^t, \gamma^t$ for textual tokens. Figure~\ref{subfig:mm_dit_block_scheme} provides a visual representation of the MM-DiT block structure, and the forward pass can be expressed as:

    \begin{equation} \label{eq:mmdit_layers}
    \begin{split}
        &x_l^v = x_{l-1}^v + \gamma_1^v MHSA(\alpha_1^v LN(x_{l-1}) + \beta_1^v), \\
        &x_l^t = x_{l-1}^t + \gamma_1^t MHSA(\alpha_1^t LN(x_{l-1}) + \beta_1^t), \\
        &x_{l+1}^v = x_{l}^v + \gamma_2^v FF(\alpha_2 LN(x_l^v) + \beta_2^v), \\
        &x_{l+1}^t = x_{l}^t + \gamma_2^t FF(\alpha_2 LN(x_l^v) + \beta_2^t),
    \end{split}
    \end{equation}
    
\noindent
where $x_l^v$ and $x_l^t$ correspond to the transformed tokens for the visual and textual modalities, respectively.

\begin{figure}[t]
\begin{center}

\begin{subfigure}{\linewidth}

\includegraphics[width=\linewidth,trim={0.5em 0.5em 0.5em 0.5em},clip]{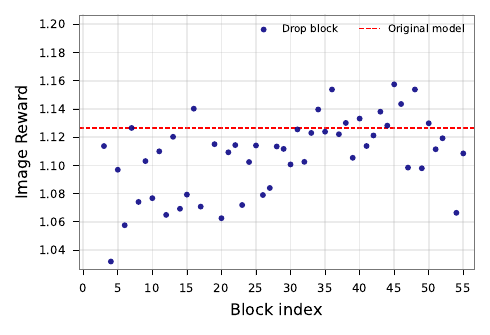}
\caption{DiT block ablation}
\label{subfig:ir_block}
\end{subfigure}
\vfill
\begin{subfigure}{\linewidth}

\includegraphics[width=\linewidth,trim={0.5em 0.5em 0.5em 0.5em},clip]{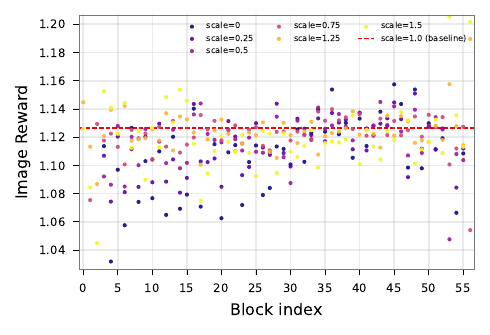}
\caption{DiT block scaling}
\label{subfig:ir_scale}
\end{subfigure}

\caption{\textbf{Motivational Experiment:} Our findings reveal that the contribution of DiT blocks is not fully optimized. We demonstrate that their performance can be enhanced through a straightforward output scaling using a scalar multiplier.}
\label{fig:motivational_experiment}

\end{center}
\end{figure}

\begin{figure*}[t]
 \begin{center}
  \includegraphics[width=\linewidth,trim={2em 1em 2em 0.5em},clip]{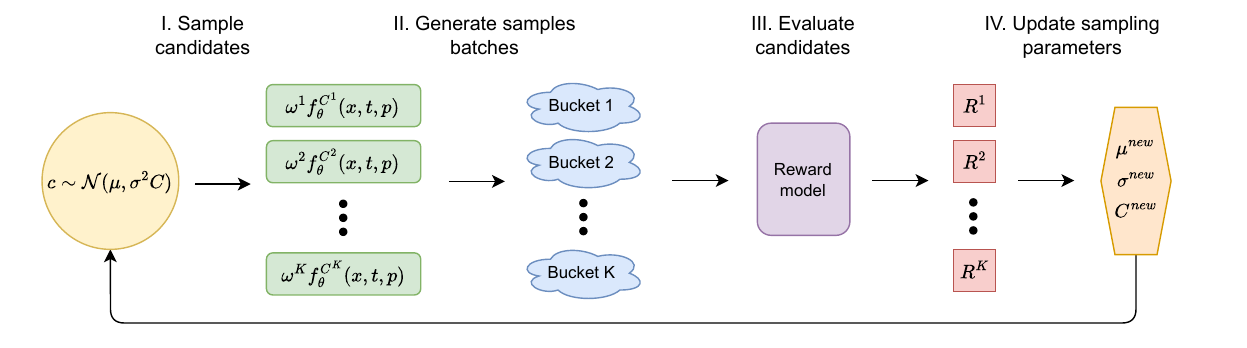}
 \end{center}
 \caption{Illustration of calibration parameter search procedure.}
 \label{fig:calibration_scheme}
\end{figure*}

\subsection{Motivation} \label{subsec:motivation}
Previous works~\cite{avrahami2025stable, wei2025freeflux} have shown that, despite the similar architectural design across DiT blocks, their contributions to the overall model performance are uneven. Notably, Stable Flow~\cite{avrahami2025stable} identified the presence of "vital layers", whose exclusion during the inference process produces significant shifts in model outputs. Motivated by these findings, we aim to explore how the exclusion of individual DiT blocks impacts the model's overall quality in generative tasks.

To systematically evaluate the importance of individual layers within DiT, we devised a structured analysis framework. Using Qwen 3~\cite{yang2025qwen3} model, we first generated a set $P$ comprising $k = 64$ diverse text prompts. These prompts were used to produce a baseline set of images, $G^{base}$ utilizing FLUX~\cite{flux2024} model. Next, for each DiT layer $l \in L$, we performed a controlled ablation by bypassing the layer output via its residual connection (i.e.  in Formula~\ref{eq:dit_layers} and ~\ref{eq:mmdit_layers}, we multiply each $\gamma$ by 0). This process produced a collection of partially ablated image sets, $\{G^l\}$, from the same text prompts. To ensure statistical validity, we repeated each experimental configuration across 5 different random seeds.

To assess the impact of the layer, we compute Image Reward~\cite{xu2023imagereward} scores for both the baseline image set $G^{base}$ and the ablated sets $\{G^l\}$. The results, presented in Figure~\ref{subfig:ir_block}, revealed an intriguing outcome: \textit{removing certain layers can occasionally enhance the quality of generated images rather than degrade it.}

Inspired by this result, we extend the experiment and generate a set of images $\{G^l_s\}$, where $s$ denotes a block output scaling coefficient, we use different $s \in \{0, 0.25, 0.5, 0.75, 1.25, 1.5\}$, $s=0$ corresponts to block ablation, $s=1$ corresponds to original model. The results, shown in Figure~\ref{subfig:ir_scale}, presented another remarkable insight: \textit{for each DiT block $l$, there exists an optimal scaling factor $s$ that improves the model's performance over its original configuration.}

\subsection{Calibri} \label{sec:calibri_description}

Based on the obtained insights, we present a simple yet effective method, named \textit{Calibri}, aimed at enhancing the generative capabilities of the diffusion transformer by calibrating only a minimal subset of the model's parameters (Figure~\ref{fig:calibration_scheme}).

\textbf{Problem formulation.} The calibration process can be formulated as an optimization problem. Let $ c \in \mathbb{R}^{L+1} $ represent the parameter vector of the diffusion transformer, where $ L+1 $ corresponds to the total number of parameters selected for calibration. The goal is to find the optimal parameter configuration $ c^* $ that maximizes the reward function:
\begin{equation} \label{eq:calibri_optimization}
    c^* = \arg \max_{c} R(c),
\end{equation}

\noindent
where $R(*)$ is a scalar-valued function measuring the performance of the diffusion transformer on the given task.

\textbf{Search space.} The search space for calibrating the model is determined by the specific locations within the diffusion transformer where adjustments are applied. For a DiT-based diffusion or flow-based model $f_{\theta}(x, t, p)$, the calibration parameters are defined as $c = \omega \cup \{s_i\}_{s=1}^{L}$, where $\omega$ denotes output-level calibration weights, and $\{s_i\}_{s=1}^{L}$ represents internal-layer calibration parameters. The calibrated model output is thus expressed as $\omega f^s_{\theta}(x, t, p)$, where the applied calibration weights refine both external outputs and internal computations.

We introduce three levels of granularity for internal-layer calibration parameters, tailored to the structural hierarchy of diffusion transformers:

\begin{enumerate}
    \item \textbf{Block Scaling:} As motivated in Section~\ref{subsec:motivation}, block-wise scaling offers a coarse calibration technique by uniformly adjusting the outputs of Attention and MLP layers within the same architectural block using a shared scaling coefficient $\mathbf{s}$.

    \item \textbf{Layer Scaling:} Extending the calibration to finer granularity, layer-wise scaling adjusts individual layers within a block using distinct coefficients. This method provides greater flexibility in refining model behavior beyond uniform block-level adjustments.

    \item \textbf{Gate Scaling:} Gate-wise calibration becomes particularly important for architectures with multimodal interactions, such as MM-DiT. Here, visual and textual tokens are processed through distinct gates, each requiring specialized calibration to optimize their interaction dynamics.

\end{enumerate}

\textbf{Calibration parameters search procedure.} To identify optimal calibration coefficients, we employ the Covariance Matrix Adaptation Evolution Strategy (CMA-ES)~\cite{hansen2003reducing, hansen2019pycma}, a powerful gradient-free optimization approach. CMA-ES optimizes an objective function $R : \mathbb{R}^d \rightarrow \mathbb{R}$ by iteratively refining a sampling distribution based on a multivariate Gaussian, $\mathcal{N}(\mu, \sigma^2 C)$, where $\mu \in \mathbb{R}^d$ represents the mean vector, $\sigma \in \mathbb{R}_{>0}$ is the step-size, and $C \in \mathbb{R}^{d \times d}$ is the covariance matrix. 

The method scheme is depicted in the Figure~\ref{fig:calibration_scheme}. At each iteration, candidate solutions are drawn from this Gaussian distribution and evaluated using the objective function. CMA-ES updates the mean vector by moving toward higher-performing candidates while adapting the covariance matrix to reflect successful directions in the search space. This iterative refinement allows efficient exploration and exploitation, optimizing calibration coefficients for improved model performance over successive iterations.

\subsection{Calibri Ensemble} \label{sec:calibri_ensemble}

\textit{Calibri}  also introduces an intriguing perspective when applied in an ensemble setting. Unlike traditional inference approaches, where combining similar models might offer negligible benefits, our method enables the calibration of an ensemble of $N$ models simultaneously. Specifically, the ensemble is represented as: 

$$ F^{\{c_i\}_{i=1}^N}(x, t, p) = \sum_{i=1}^N \omega_i f^{s_i}_{\theta}(x, t, p | \emptyset),$$ 

\noindent
where $\omega_i$ denotes the weight assigned to the $i$-th model, $f^{s_i}_{\theta}$ represents the calibrated with internal-layer calibration parameters $s_i$, and $\{x, t, p\}$ are the input signals, time step, and additional conditioning inputs, respectively.

In this case, the optimization problem~\ref{eq:calibri_optimization} is reformulated:

\begin{equation} \label{eq:calibri_ensemble_optimization}
    \{c_i^*\}_{i=1}^N = \arg \max_{\{c_i\}_{i=1}^N} R(\{c_i\}_{i=1}^N),
\end{equation}
\noindent
where $R(\{c_i\}_{i=1}^N)$ evaluates the overall performance of the ensemble. This ensemble calibration allows us to leverage the diversity among $N$ optimized models, resulting in enhanced generative performance and robustness.

\textbf{Relation to Classifier-Free Guidance.} \textit{Calibri ensemble} framework seamlessly integrates into the Classifier-Free Guidance paradigm. In this specific case, the ensemble size is $N=2$, and optimization is performed following Problem~\ref{eq:calibri_ensemble_optimization}. By calibrating two models representing distinct guiding roles (conditional and unconditional), \textit{Calibri} enhances generation while maintaining diversity and precision.

%% file: sec/4_experiments.tex
\section{Experiments}
\label{sec:experiments}

% We conducted a number of  experiments to validate the effectiveness of  \textit{Calibri} across multiple text-to-image  generation models and  evaluated our approach using diverse metrics and compared against both baseline models and models fine-tuned with recent alignment methods.

\textbf{Baselines:} To evaluate the effectiveness of \textit{Calibri}, we compare its performance across several state-of-the-art, open-source DiT-based text-to-image models. Specifically, we conduct experiments on FLUX.1-dev~\cite{flux2024}, Stable Diffusion 3.5 Medium (SD-3.5M)~\cite{esser2024scaling}, and Qwen-Image~\cite{wu2025qwen}, all of which represent the cutting edge in text-to-image generation. Additionally, we test \textit{Calibri} using the SD-3.5M model checkpoint fine-tuned with Flow-GRPO~\cite{liu2025flow} to analyze its performance in alignment-sensitive setups.

\textbf{Implementation Details:} For our experiments, we used train and test prompts from T2I-compbench++~\cite{t2icompbenchplus}: train prompts were used to sample buckets for candidate evaluation in the CMA-ES algorithm and test prompts were used for intermediate reward evaluations to select the best coefficients. We used HPSv3~\cite{ma2025hpsv3} to track image preference and Q-Align ~\cite{qalign} to track image quality during training. For hyperparameters of CMA-ES we used commonly used parameters: the initial sigma was set to 0.25, the number of candidates was set to $4 + \lfloor 3 \ln d \rfloor \approx 20 \pm 3$ for considered models, where $d$ is the dimension of the search space and represents the number of coefficients to be tuned. We have fixed the bucket size to 16, the image resolution to 512 and the number of inference steps to 15 for training -- we found it is the lowest number of steps that achieve satisfactory generation quality across several models.

\textbf{Evaluation and Metrics:}  We used HPDv3 test prompts for evaluation. To measure the final metrics, we used HPSv3~\cite{ma2025hpsv3}, Q-Align~\cite{qalign} and ImageReward~\cite{xu2023imagereward}.

% and Pickscore~\cite{kirstain2023pick}.

\begin{table}[t]
\centering
\caption{Comparison of various granularity levels for internal-layer calibration applied to the Flux model. All experiments are conducted using the Flux model, with evaluation performed on the HPSv3 test set.}
\label{tab:scaling_ablation}
% \resizebox{\columnwidth}{!}{%
\begin{tabular}{lccccc}
\toprule
\textbf{Scaling}& \textbf{N params}  & \textbf{iters} & \textbf{HPSv3} & \textbf{IR} & \textbf{Q-Align}\\ \hline
 --       & --    & --  & 11.41          & 1.15        & 4.85            \\
 Block          & 57 & 200 & 13.29          & 1.17        & \textbf{4.91}             \\
 Layer          & 76  & 410 & \underline{13.41}          & \textbf{1.24}        & \underline{4.90}             \\
 Gate           & 114  & 960 & \textbf{13.48}          & 1.18        & 4.88            \\

\bottomrule
\end{tabular}%
% }
\end{table}

\begin{figure}[t]
  \centering
  \includegraphics[width=\linewidth]{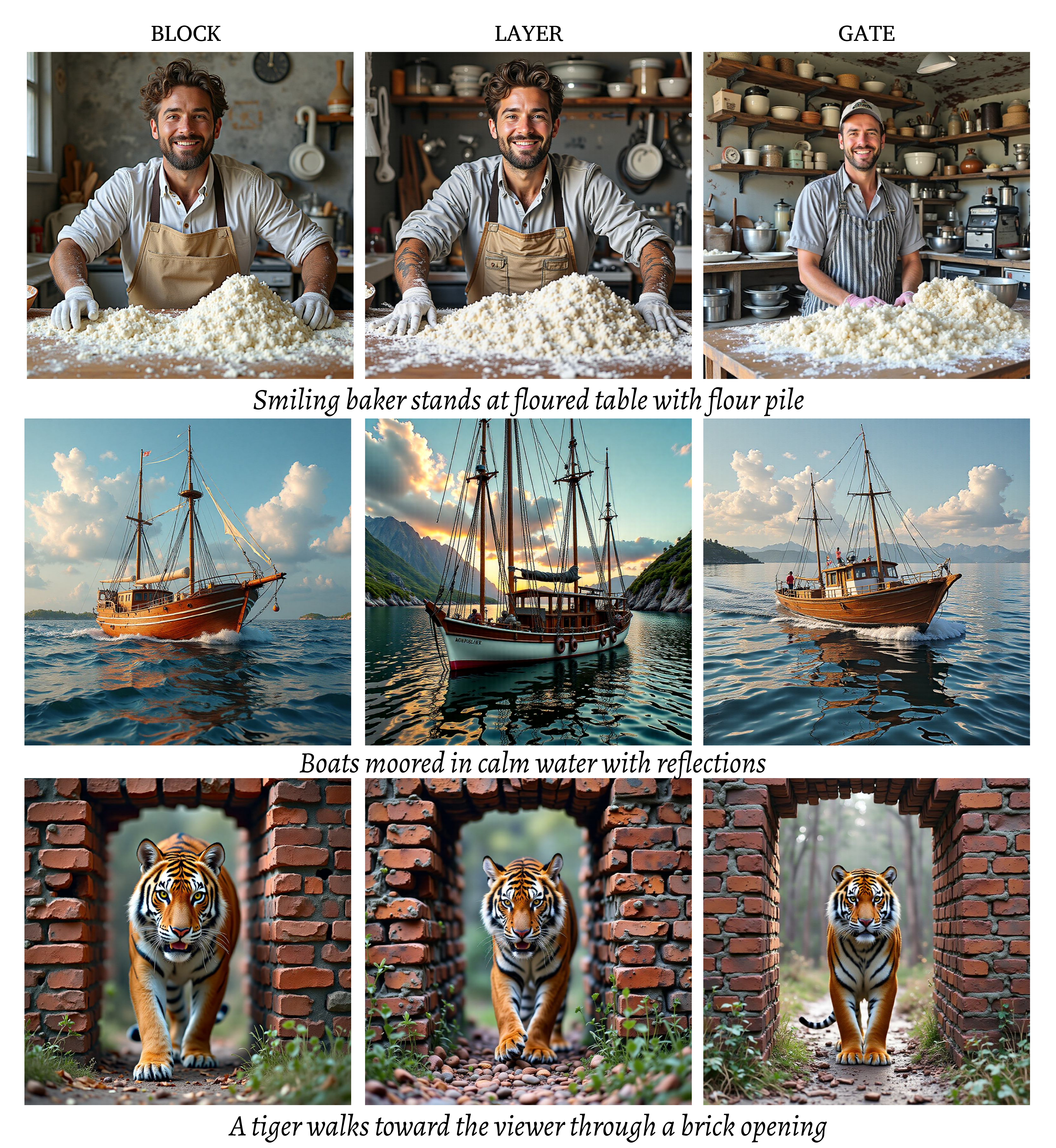}
  \caption{Quantitative comparison of various granularity levels for internal-layer calibration applied to the Flux model.}
  \label{fig:scaling_ablation}
\end{figure}

\subsection{Calibri Design Decisions}

\textbf{Search space.} To evaluate the effectiveness of scaling granularity, we consider three options introduced in Section~\ref{sec:calibri_description}: block scaling, layer scaling, and gate scaling. These scaling methods progressively increase the number of parameters available for optimization, as detailed in Table~\ref{tab:scaling_ablation}. All experiments are conducted using \textit{Calibri} applied to the Flux model, with optimization guided by HPSv3 reward.

While gate scaling achieves the highest value of the target reward (HPSv3), it underperforms on several alternative rewards. In contrast, layer scaling yields more consistent improvements across multiple reward functions, and Figure~\ref{fig:scaling_ablation} illustrates its advantage over the other scaling methods. Overall, the resulting performance across the three schemes is relatively similar, but their training speeds differ substantially, which is an important factor when choosing the appropriate scaling strategy.

\begin{figure}[t]
  \centering
  \includegraphics[width=\linewidth]{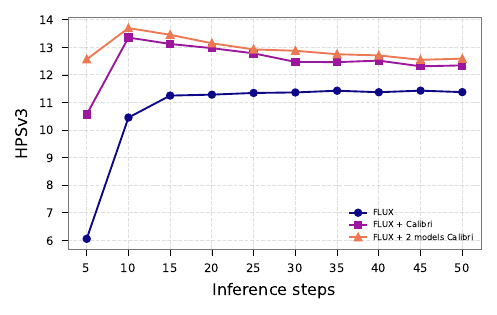}
  \caption{Comparison between \textit{Calibri Ensemble} and original model across several inference steps.}
  \label{fig:hpsv3_vs_nfe}
\end{figure}

\textbf{N models.}
The \textit{Calibri Ensemble} method (Section~\ref{sec:calibri_ensemble}) allows us to aggregate multiple differently calibrated models into a single sampler.
To validate this approach, we evaluate \textit{Calibri Ensemble} on FLUX guided by HPSv3 reward with \(N \in \{1, 2\}\) models using HPDv3 prompts.
In the experiments, we use block scaling, as it empirically yields the fastest convergence.
Since FLUX is a CFG-distilled model, we pass the same prompt to each model instance and then combine their contributions within the \textit{Calibri} framework.

We also note that for \(N = 2\) with block scaling, the \textit{Calibri Ensembling} method generalizes Skip Layer Guidance (also referred to as Spatiotemporal Guidance~\cite{stgguidance}), which can be seen as a special training-free case of Autoguidance~\cite{autoguidance}.
The results show that ensembling calibrated models consistently increases the HPSv3 reward across all inference steps, as illustrated in Figure~\ref{fig:hpsv3_vs_nfe}.

\textbf{NFE.}
Another notable observation in Figure~\ref{fig:hpsv3_vs_nfe} is that \textit{Calibri Ensembling} shifts the optimal number of sampling steps from 30--50 in the baseline to only 10--15 steps.
This substantially reduces the number of function evaluations required to achieve strong performance, making inference both faster and more computationally efficient.

\subsection{Different Backbones}

\begin{figure*}[t]
 \begin{center}
  \includegraphics[width=\textwidth,trim={10em 3em 10em 0em},clip]{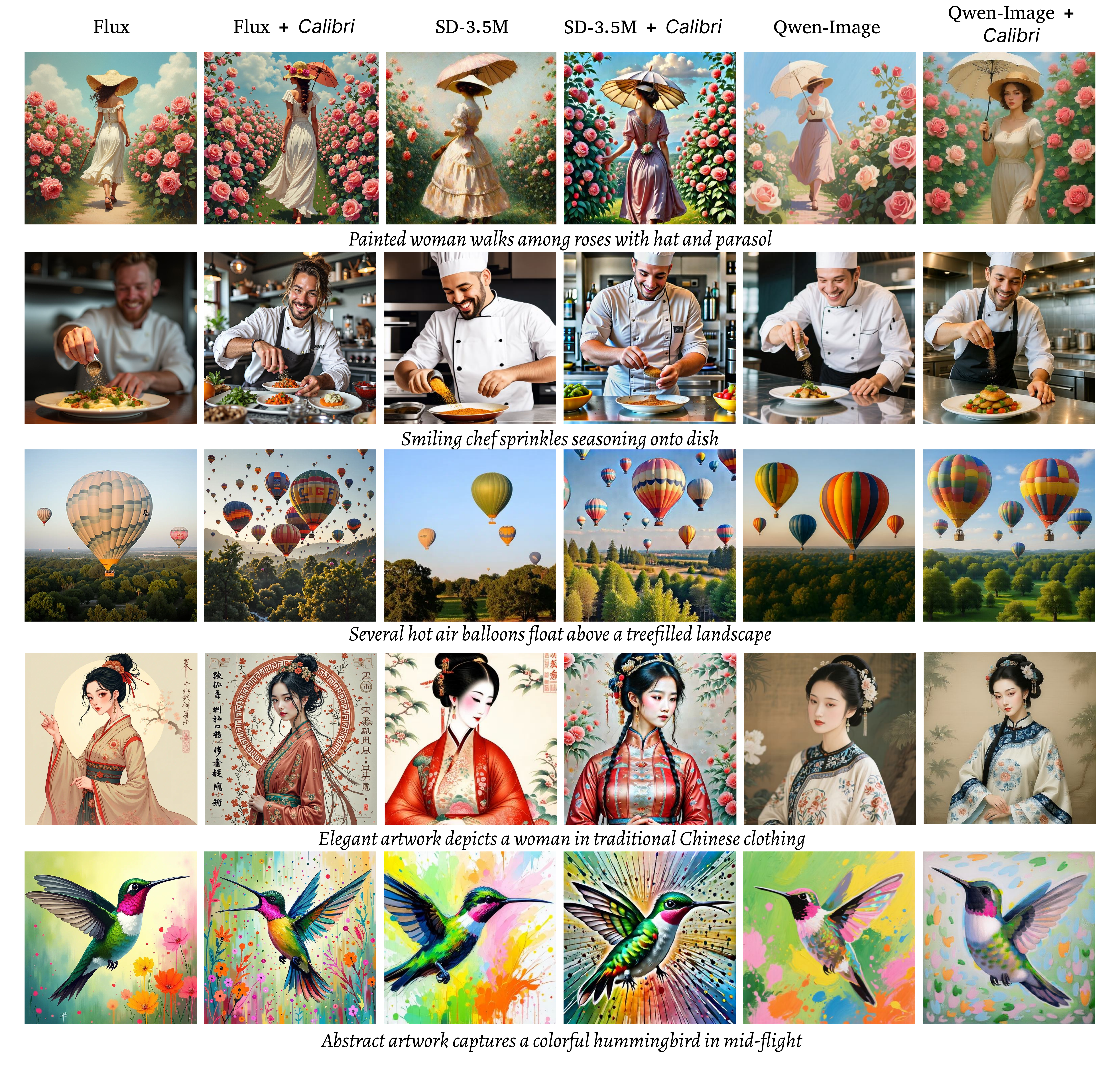}
 \end{center}
 \caption{Qualitative evaluation of generation quality improvements across various baseline models. Models have same NFE as in Table~\ref{tab:main_results}.}
 \label{fig:qualitative_main}
\end{figure*}

We evaluate \textit{Calibri} on three representative T2I models: Flux, SD-3.5M, and Qwen-Image. Quantitative results, presented in Table~\ref{tab:main_results}, demonstrate consistent performance improvements across all baseline models when using \textit{Calibri}. Notably, \textit{Calibri} achieves these enhanced metrics while requiring significantly fewer inference steps -- 15 steps compared to 30 for Flux, 40 for SD-3.5M, and 50 for Qwen-Image. Furthermore, qualitative comparisons in Figure~\ref{fig:qualitative_main} illustrate superior output quality of \textit{Calibri}, reinforcing its effectiveness and practical advantages.

\begin{table}[t]
\centering
\caption{Quantitative evaluation of generation quality improvements across various baseline models. Notably, \textit{Calibri} achieves superior metric scores while requiring fewer inference steps.}
\label{tab:main_results}
\resizebox{\columnwidth}{!}{%
\begin{tabular}{lccccc}
\toprule
\textbf{Model}              & \textit{\textbf{Calibri}} & \textbf{HPSv3} & \textbf{IR} & \textbf{Q-Align} & \textbf{NFE} \\ \hline
\multirow{2}{*}{FLUX}       &  \xmark                         & 11.41          & 1.15                  & 4.85             & 30              \\
                            &  \cmark                         & \textbf{13.48}          & \textbf{1.18}                  & \textbf{4.88}             & 15              \\ \hline
\multirow{2}{*}{SD-3.5M}    &  \xmark                          & 11.15          & 1.10                   & 4.74             & 80              \\
                            &  \cmark                          & \textbf{14.10}          & \textbf{1.17}                  & \textbf{4.91}             & 30              \\ \hline
\multirow{2}{*}{Qwen Image} &   \xmark                        & 11.26          & 1.16                  & 4.55             & 100              \\
                            &   \cmark                         & \textbf{12.95}          & \textbf{1.18}                  & \textbf{4.73}             & 30              \\ 
\bottomrule
\end{tabular}
}
\end{table}

To verify genuine improvements beyond reward metrics, we conducted a large-scale user study (200 users, 5,600 assessments, 150 HPDv3 test set prompts) on Flux.1-dev and Qwen-Image. Table~\ref{tab:user_study} shows evaluators decisively prefer \textit{Calibri} in both Overall Preference and Text Alignment, confirming genuine perceptual gains (not reward artifacts). Calibrated models are also 2--3.3$\times$ faster than baselines.

\begin{table}[t]
\centering
\caption{Human evaluation: \textit{Calibri} vs. Baselines win rates \%.}
\label{tab:user_study}

\resizebox{\linewidth}{!}{
\begin{tabular}{lcccccc}
\toprule
\multirow{2}{*}{Methods} & \multicolumn{3}{c}{Overall Preference} & \multicolumn{3}{c}{Text Alignment}  \\  
                         & \textit{Calibri}       & Equal       & Original        & \textit{Calibri}   & Equal         & Original           \\ \hline
\textit{Flux}              & \textbf{51.87}   &  7.33   & 40.80               & \textbf{38.71}  & 37.68  & 23.61            \\
\textit{Qwen-Image}          & \textbf{54.62}   &  7.91   & 37.47               & \textbf{40.29}  & 37.65  & 22.06  \\

\bottomrule
\end{tabular}
}

\end{table}

\subsection{Combining Calibri with Alignment Methods}

\begin{table}[t]
\centering
\caption{Comparison of \textit{Calibri} and Flow-GRPO~\cite{liu2025flow} on SD-3.5M. \textit{Calibri} achieves comparable performance with $10^5$ fewer parameters and can be combined with alignment methods to boost either the same or different target metrics.}
\label{tab:alignment_results}
\resizebox{\columnwidth}{!}{%
\begin{tabular}{lccccc}
\toprule
\textbf{Flow-GRPO}         & \textbf{Calibri} & \textbf{HPSv3} & \textbf{PickScore} & \textbf{Q-Align} & \textbf{NFE} \\ \hline
\multirow{2}{*}{\xmark}        & \xmark             & 11.15          & 22.40              & 4.74             & 80           \\
                           & PickScore        & \textbf{12.47}          & \textbf{23.13}              & \textbf{4.91}             & 30           \\ \hline
\multirow{2}{*}{PickScore} & \xmark             & 12.67          & 23.78              & \textbf{4.92}             & 80           \\
                           & PickScore        & \textbf{12.96}          & \textbf{23.93}              & 4.85             & 30           \\ \hline
\multirow{2}{*}{GenEval}   & \xmark             & 10.16          & 22.22              & 4.69             & 80           \\
                           & HPSv3            & \textbf{14.18}         & 22.22              & \textbf{4.88}             & 30   \\
\bottomrule
\end{tabular}%
}
\end{table}

\begin{figure*}[t]
 \begin{center}
  \includegraphics[width=\linewidth,trim={6em 10em 8em 3em},clip]{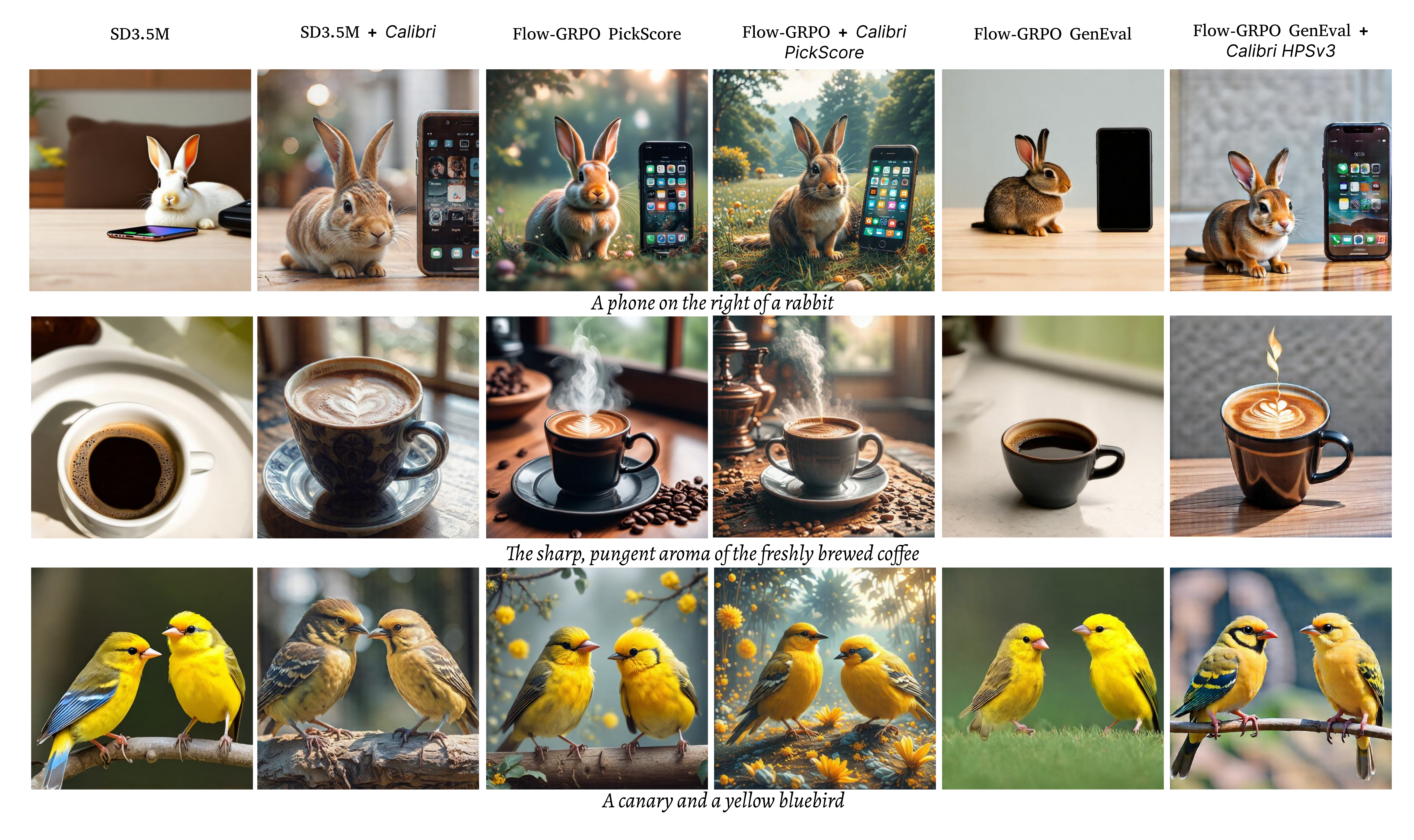}
 \end{center}
 \caption{Qualitative comparison of \textit{Calibri} and Flow-GRPO~\cite{liu2025flow} on SD-3.5M. \textit{Calibri} achieves comparable performance with $10^5$ fewer parameters and can be combined with alignment methods to boost either the same or different target metrics.}
 \label{fig:alignment_results}
\end{figure*}

We assess the effectiveness of \textit{Calibri} in combination with Alignment methods on three distinct SD-3.5M checkpoints: the pretrained base model, Flow-GRPO~\cite{liu2025flow} checkpoint aligned with PickScore~\cite{kirstain2023pick} reward, and Flow-GRPO~\cite{liu2025flow} checkpoint aligned with GenEval~\cite{ghosh2023geneval} metric. Table~\ref{tab:alignment_results} and Figure~\ref{fig:alignment_results} present results for three experimental setups.

First, we examine the impact of applying \textit{Calibri} to the base model for optimizing PickScore. Notably, the procedure improves target metric PickScore as well as HPSv3 and Q-Align, indicating a broader positive impact on model performance. Additionally, the model optimized for PickScore using \textit{Calibri} achieves comparable results to the optimization achieved by Flow-GRPO, despite \textit{Calibri} updating only 216 parameters compared to the 18.78M parameters updated by Flow-GRPO.

Next, we investigate the application of \textit{Calibri} to a Flow-GRPO checkpoint that was already optimized for PickScore. Our results show that \textit{Calibri} further improves performance improvements of the model, showing its utility in enhancing models already aligned to a specific reward.

Finally, we apply \textit{Calibri} to a Flow-GRPO checkpoint trained to maximize the GenEval metric. As demonstrated in Table~\ref{tab:alignment_results} and Figure~\ref{fig:alignment_results}, \textit{Calibri} integrates efficiently with standard alignment methods and significantly boosts metric performance across the board. These findings highlight \textit{Calibri}'s versatility and effectiveness in improving model alignment with various optimization targets.

\subsection{Calibration Cost} \label{sec:calibration_cost}

We report calibration cost in NVIDIA H100 GPU‑hours in Table~\ref{tab:calibration_cost}. Convergence depends on both search‑space size and pre‑trained model quality: larger search spaces (more parameters) generally require more calibration iterations, while higher‑quality models tend to converge faster. In our experiments, stronger models (Flux, Qwen) converged in roughly 200–960 iterations, whereas a weaker model (SD‑3.5M) required about 2,280 iterations. The total calibration cost therefore ranges from 32 to 356 H100 GPU‑hours. Crucially, this is a one‑time, offline cost — for example, calibrating Flux (Block) takes only 32 H100 GPU‑hours and yields an approximately 2$\times$ permanent speed‑up at inference.

\begin{table}[t]
\centering
\caption{Calibration cost.}
\label{tab:calibration_cost}
\resizebox{\linewidth}{!}{
\begin{tabular}{lccccc}
\toprule
Model        & \multicolumn{3}{c}{Flux}                                                  & SD-3.5M              & Qwen-Image           \\
Scaling type & Block                  & Layer                  & Gate                    & Gate                 & Gate                 \\ \hline
CFG type     & \multicolumn{3}{c}{CFG Distill}                                           & \multicolumn{2}{c}{Standard CFG}                \\
N params     & 57 & 76 & 114 & 216 & 482 \\
N iters      & 200                    & 410                    & 960                     & 2280                 & 630                  \\
GPU hours    & 32                     & 64                     & 150                     & 356                & 286 \\   
\bottomrule
\end{tabular}
}
\end{table}

%% file: sec/5_conclusion.tex
\section{Conclusion}
\label{sec:conclusion}

In this work, we introduced \textit{Calibri}, a novel and parameter-efficient approach to enhance the generative capabilities of Diffusion Transformers (DiTs). By uncovering the potential of a single learned scaling parameter to optimize the contributions of DiT components, we demonstrated that significant performance improvements can be achieved with minimal parameter modifications. Framing the calibration process as a black-box optimization problem solved via the CMA-ES evolutionary strategy, \textit{Calibri} adjusts only $\sim 10^2$ parameters while delivering consistently improved generation quality. Additionally, the proposed inference-time scaling technique, \textit{Calibri Ensemble}, effectively combines calibrated models to further enhance results.

Our extensive empirical evaluation across a range of text-to-image diffusion models confirmed the effectiveness and efficiency of \textit{Calibri}, highlighting its ability to achieve superior generative quality with reduced computational costs. Notably, \textit{Calibri} successfully reduces the number of inference steps required for image generation while retaining high-quality outputs, making it a practical solution for real-world applications where computational efficiency is critical.

% Future work will explore extending \textit{Calibri} to other generative architectures, as well as investigating its broader applicability to tasks beyond text-to-image synthesis. By providing a lightweight yet powerful framework for optimizing diffusion-based generative models, we hope \textit{Calibri} inspires further innovation in efficient model calibration and scalable generative design.

\section*{Acknowledgments}
We thank Vera Soboleva for preparing the paper’s illustrations and carefully proofreading the manuscript, and Alexander Gambashidze for his insightful discussions.

This work was supported by the The Ministry of Economic Development of the Russian Federation in accordance with the subsidy agreement (agreement identifier 000000C313925P4H0002; grant No 139-15-2025-012).

%% file: sec/X_suppl.tex
\clearpage
\setcounter{page}{1}
\maketitlesupplementary

% \section{Rationale}
% \label{sec:rationale}
% % 
% Having the supplementary compiled together with the main paper means that:
% % 
% \begin{itemize}
% \item The supplementary can back-reference sections of the main paper, for example, we can refer to \cref{sec:intro};
% \item The main paper can forward reference sub-sections within the supplementary explicitly (e.g. referring to a particular experiment); 
% \item When submitted to arXiv, the supplementary will already included at the end of the paper.
% \end{itemize}
% % 
% To split the supplementary pages from the main paper, you can use \href{https://support.apple.com/en-ca/guide/preview/prvw11793/mac#:~:text=Delete%20a%20page%20from%20a,or%20choose%20Edit%20%3E%20Delete).}{Preview (on macOS)}, \href{https://www.adobe.com/acrobat/how-to/delete-pages-from-pdf.html#:~:text=Choose%20%E2%80%9CTools%E2%80%9D%20%3E%20%E2%80%9COrganize,or%20pages%20from%20the%20file.}{Adobe Acrobat} (on all OSs), as well as \href{https://superuser.com/questions/517986/is-it-possible-to-delete-some-pages-of-a-pdf-document}{command line tools}.

\section*{Supplementary Material Structure}

This supplementary document is organized as follows:

\begin{enumerate}
    \item Section~\ref{sec:limitations} elaborates on the limitations of the proposed methodology, providing a detailed analysis.

    \item Section~\ref{sec:gen_diversity} analyzes the diversity of images generated by the method, both before and after incorporating the Calibri technique.

    \item Section~\ref{sec:different_rewards} explains the rationale behind the chosen reward model, highlighting its impact on the system's performance.
    \item Section~\ref{sec:different_parameter_search} discusses the motivation for using the CMA-ES approach as the parameter search method, justifying its effectiveness.
\end{enumerate}

\section{Limitations} \label{sec:limitations}

Our calibration coefficients selection method, \textit{Calibri}, leverages a reward model~\cite{ma2025hpsv3} as its objective function. Reward models are trained to approximate user preferences for generated images, which enables \textit{Calibri} to optimize the selection process effectively.

\begin{figure}[h]
  \centering
  \includegraphics[width=\linewidth]{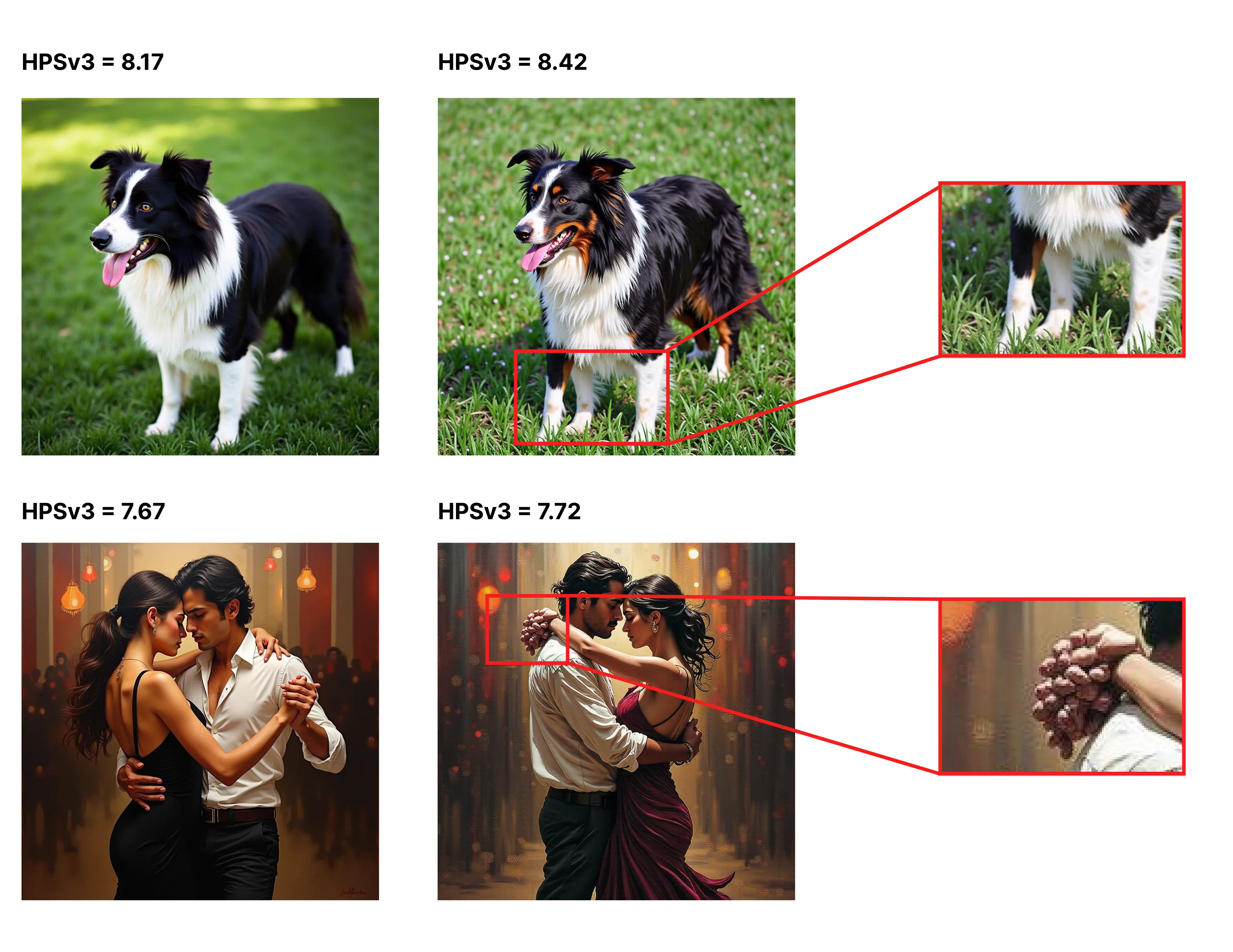}
  \caption{Limitations of modern reward models.}
  \label{fig:limitations}
\end{figure}

However, despite substantial advancements in reward modeling in recent years, current reward models still exhibit notable limitations. Specifically, they often demonstrate insufficient sensitivity to generation artifacts such as anatomical inconsistencies—examples include extra limbs, distorted fingers, and other visually unrealistic features, as illustrated in Figure~\ref{fig:limitations}.

These shortcomings in reward models can impact the performance of \textit{Calibri}, resulting in the selection of a suboptimal set of coefficients. Addressing these limitations is crucial for further improving the robustness and overall effectiveness of the calibration process.

We anticipate that ongoing advancements in reward modeling techniques will mitigate these issues and significantly enhance their sensitivity to such artifacts, ultimately improving the performance of \textit{Calibri} in future iterations.

\section{Generation diversity} \label{sec:gen_diversity}

Optimizing diffusion models using reward models often leads to a reduction in generation diversity, as highlighted by recent findings~\cite{sorokin2025imagerefl}. Since \textit{Calibri} employs a reward model as its optimization objective, it is important to evaluate how this approach affects generation diversity. In Table~\ref{tab:diversity_results}, we present a comparison of generation diversity between the original model (SD-3.5M)~\cite{esser2024scaling} and models optimized by \textit{Calibri} and Flow-GRPO~\cite{liu2025flow}.

The results demonstrate that the generation diversity of SD-3.5M with 40 inference timesteps remains comparable to that of the model optimized by \textit{Calibri}, which achieves comparable diversity while requiring only 15 inference timesteps. Importantly, despite its reduced inference time, the model optimized by \textit{Calibri} exhibits significantly higher generation quality compared to the original model. 

In contrast, Flow-GRPO reduces generation diversity from 0.20 to 0.15 and fails to accelerate inference time. Furthermore, when \textit{Calibri} is applied to the model already optimized by Flow-GRPO, it does not introduce any further changes in generation diversity. This result underscores the efficiency and robustness of \textit{Calibri} in preserving diversity while enhancing generation quality and reducing inference time.

\begin{table}[h]
\centering
\caption{Comparison of Generation Diversity for SD-3.5M~\cite{esser2024scaling}  Optimized by \textit{Calibri} and Flow-GRPO~\cite{liu2025flow}.}
\label{tab:diversity_results}
\resizebox{\columnwidth}{!}{%
\begin{tabular}{ccccccc}
\toprule
\textbf{Flow-GRPO}         & \textbf{Calibri} & \textbf{HPSv3} & \textbf{PickScore} & \textbf{Q-Align} & \textbf{Dino Diversity} & \textbf{NFE}  \\ \hline
\multirow{3}{*}{\xmark}        & \xmark             & 11.15          & 22.40              & 4.74             & $0.20 \pm \text{0.06}$ & 80           \\
 & \xmark             &  9.5              & 22.04    & 4.51     &  $ 0.25 \pm \text{0.08}$              & 30     \\

                           & PickScore        & \textbf{12.47} & \textbf{23.13}      & \textbf{4.91}    &  $0.20 \pm \text{0.06}$                   & 30           \\ \hline
\multirow{3}{*}{PickScore} & \xmark             & 12.67          & 23.78              & \textbf{4.92}    &  $0.15 \pm \text{0.06}$                  & 80           \\
 & \xmark             &  12.514        & 23.76      & 4.91  &  $ 0.15 \pm \text{0.05} $                 & 30     \\
                           & PickScore        & \textbf{12.96} & \textbf{23.93}      & 4.85             &  $0.15   \pm \text{0.05}$                  & 30           \\ 
\bottomrule
\end{tabular}%
}
\end{table}

\section{Different Rewards} \label{sec:different_rewards} 

\begin{figure}[h]
  \centering
  \includegraphics[width=\linewidth]{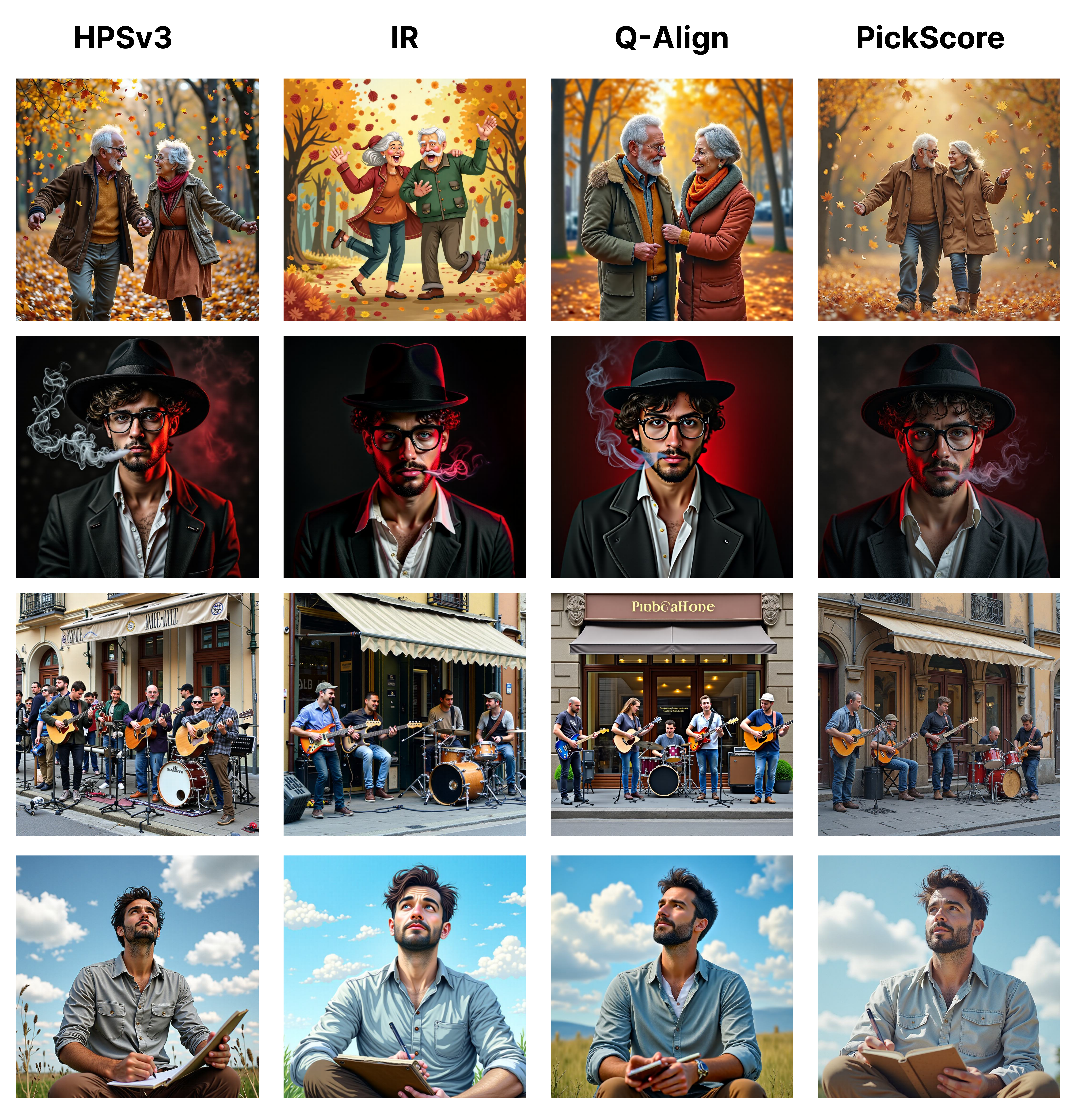}
  \caption{Illustration of \textit{Calibri} with different rewards as objective.}
  \label{fig:different_rewards}
\end{figure}

\begin{table}[h]
\centering
\caption{Quantitative comparison of \textit{Calibri} layer scale on Flux~\cite{flux2024} across different reward models.}
\label{tab:calibri_rewards}
\resizebox{\columnwidth}{!}{%
\begin{tabular}{cccccc}
\toprule
\textbf{Calibri} & \textbf{HPSv3} & \textbf{IR} & \textbf{Q-Align} & \textbf{PickScore} & \textbf{NFE} \\
\midrule
\xmark      & 11.41 & 1.15 & 4.85 & 22.88 & 30 \\
HPSv3       & \textbf{13.41} & \textbf{1.24} & \textbf{4.90}  & \underline{23.07} & 15 \\
ImageReward & 11.06 & 1.17 & 4.70  & 22.47 & 15 \\
Q-Align     & 11.65 & 1.0  & 4.89 & 22.36 & 15 \\
PickScore   & \underline{13.34} & \underline{1.2}  & \underline{4.89} & \textbf{23.24} & 15 \\
\bottomrule
\end{tabular}%
}
\end{table}

To evaluate the performance of \textit{Calibri} across different objectives, we conducted experiments using various reward models as optimization objectives. The results are summarized in Table~\ref{tab:calibri_rewards} and visually presented in Figure~\ref{fig:different_rewards}.

Calibration using the HPSv3~\cite{ma2025hpsv3} reward model achieved the most significant quality improvement across all metrics, while the Pickscore~\cite{kirstain2023pick} reward exhibited similarly strong performance. Notably, we observed that calibrating with the most effective reward model not only enhances the target metric but also yields substantial improvements across other metrics. This indicates that \textit{Calibri} is not designed as a reward hacking method tailored to specific objectives, but rather as a general-purpose technique for improving overall generation quality.

% We find other rewards are less effective for calibrating the model.

\section{CMA-ES vs gradient-based parameter search} \label{sec:different_parameter_search}

\begin{figure}[t]
  \centering
  \includegraphics[width=\linewidth]{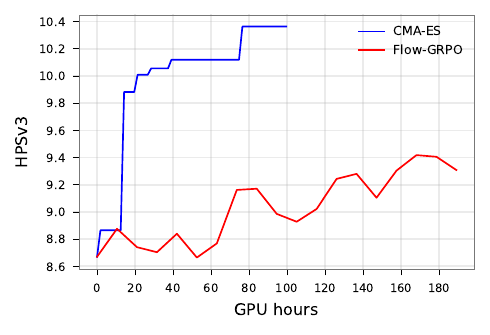}
  \caption{Comparison of CMA-ES and Flow-GRPO performance in optimizing \textit{Calibri} coefficients.}
  \label{fig:cmaes_vs_grpo}
\end{figure}

% Gradient-based algorithms are commonly used for efficient parameter search, but for diffusion models, it is complicated to use one of them because of repeated model inference steps during generation. Recent work Flow-GRPO~\cite{liu2025flow} makes a big step towards training diffusion model from reward by reformulating diffusion process as MDP. In this section, we compare CMA-ES with Flow-GRPO algorithm and provide some insights about CMA-ES training procedure. 

% To compare CMA-ES with Flow-GRPO we trained Calibri with layer scale on FLUX using these two optimizers and followed the main experiment setup with evaluation during training on T2I-compbench++~\cite{t2icompbenchplus} test prompts. For Flow-GRPO we followed the default hyperparameters configuration setup for FLUX. The evaluation curves obtained during training are presented in Figure~\ref{fig:cmaes_vs_grpo}, which demostrate that CMA-ES is much more efficient than Flow-GRPO.

% We also provide insights about the CMA-ES training procedure with Figure~\ref{fig:training_details}. CMA-ES indicates \textit{Calibri} coefficients are optimized and algorithms can be stopped when the sigma goes to a plateau and the training curve stops increasing.

Gradient-based algorithms are commonly employed for alignment of diffusion models via reward maximization. However, their application to diffusion models presents challenges due to the incompatibility of reward models with noisy latent spaces and generated images, necessitating repeated inference steps during the generation process for accurate reward computation. A recent advancement, Flow-GRPO~\cite{liu2025flow}, addresses this issue by reframing the diffusion process as a Markov Decision Process (MDP), enabling reward-driven training in such models. In this section, we evaluate the performance of CMA-ES and compare it to Flow-GRPO, while also providing additional analysis of CMA-ES training dynamics.

To compare CMA-ES with Flow-GRPO, we trained \textit{Calibri} with layer scale on FLUX using these two optimizers, following the main experimental setup with evaluation during training on the T2I-Compbench++~\cite{t2icompbenchplus} test prompts. For Flow-GRPO, we adopted the default hyperparameter configuration provided for Flux~\cite{flux2024}. The evaluation curves obtained during training are shown in Figure~\ref{fig:cmaes_vs_grpo} and demonstrate that CMA-ES is substantially more efficient than Flow-GRPO.

\begin{figure}[t]
    \centering

    \begin{subfigure}{\linewidth}
        \includegraphics[width=\linewidth,
            trim={0.5em 0.5em 0.5em 0.5em},clip]{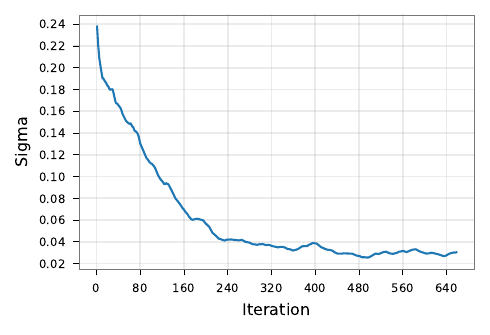}
        \caption{Sigma decrease during training and indicates when the training can be stopped.}
        \label{subfig:sigma_curve}
    \end{subfigure}

    \vspace{0.5em} 
    
    \begin{subfigure}{\linewidth}
        \includegraphics[width=\linewidth,
            trim={0.5em 0.5em 0.5em 0.5em},clip]{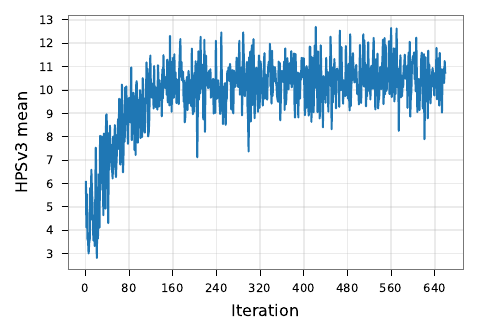}
        \caption{Training curve with the mean reward on buckets.}
        \label{subfig:train_curve}
    \end{subfigure}

    \caption{CMA-ES algorithm optimizes \textit{Calibri} coefficients for layer scale FLUX.}
    \label{fig:training_details}
\end{figure}

Additionally, we present detailed insights into CMA-ES training dynamics in Figure~\ref{fig:training_details}. Our analysis indicates that the \textit{Calibri} coefficients converge effectively during training, with optimization reaching a plateau as evidenced by the stabilization of sigma and the stagnation of improvements in the training curve. These observations suggest that training with CMA-ES can be terminated once this convergence behavior is observed, optimizing computational resources without compromising performance.

